\documentclass[conference]{IEEEtran}
\usepackage{amsmath,amssymb,graphicx,epstopdf,cite,subfigure}
\usepackage{psfrag}
\usepackage{amssymb}
\usepackage{amsmath}
\usepackage{bbm}
\usepackage{dsfont}
\usepackage{pifont}
\usepackage{cite}
\usepackage{graphics}
\usepackage{graphicx}
\usepackage{epsfig}
\usepackage{subfigure}
\usepackage{amscd}
\usepackage{accents}
\usepackage{algorithm,algpseudocode}
\newcommand{\RN}[1]{%
  \textup{\uppercase\expandafter{\romannumeral#1}}%
}

\begin{document}

\title{
Autonomous Braking System via \\
 Deep Reinforcement Learning
}
\author{\IEEEauthorblockN{Hyunmin Chae, Chang Mook Kang, ByeoungDo Kim, Jaekyum Kim, Chung Choo Chung, and Jun Won Choi}

\IEEEauthorblockA{Hanyang University, Seoul, Korea\\
Email: {hmchae@spo.hanyang.ac.kr, kcm0728@hanyang.ac.kr, \{bdkim, jkkim\}@spo.hanyang.ac.kr,} \\ {\{cchung ,junwchoi\}@hanyang.ac.kr}}
}
\maketitle

\begin{abstract}
In this paper, we propose a new autonomous braking system based on deep reinforcement learning.
The proposed autonomous braking system automatically decides whether to apply the brake at each time step when confronting the risk of collision using the information on the obstacle obtained by the sensors.
The problem of designing brake control is formulated as searching for the optimal policy in Markov decision process (MDP) model where the state is given by the relative position of the obstacle and the vehicle's speed, and the action space is defined as the set of the brake actions including 1) no braking, 2) weak, 3) mid, 4) strong brakiong actions. The policy used for brake control is learned through computer simulations using the deep reinforcement learning method called deep Q-network (DQN).
In order to derive desirable braking policy, we propose the reward function which balances the damage imposed to the obstacle in case of accident and the reward achieved when the vehicle runs out of risk as soon as possible. DQN is trained for the scenario where a vehicle is encountered with a pedestrian crossing the urban road. Experiments show that the control agent exhibits desirable control behavior and avoids collision without any mistake  in various uncertain environments.
\end{abstract}

\section{INTRODUCTION}
Safety is one of top priorities that should be pursued in realizing fully autonomous driving vehicles. For safe autonomous driving, autonomous vehicles should perceive the environments using the sensors and control the vehicle to travel to the destination without any accidents.
Since it is inevitable for an autonomous vehicle to encounter with unexpected and risky situations, it is critical to develop the reliable autonomous control systems that can cope well with such uncertainties.
Recently, several safety systems including  collision avoidance, pedestrian detection, and front collision warning (FCW) have been proposed to enhance the safety of the autonomous vehicle \cite{caacc,efcwca,pca}.

One critical component for enabling  safe autonomous driving is the autonomous braking systems which can reduce the velocity of the vehicle automatically  when a threatening obstacle is detected.  
%
%
The autonomous braking should offer  safe and comfortable brake control without exhibiting too early or too late braking.
Most  conventional autonomous braking systems are rule-based, which designate the specific brake control protocol for each different situation.
Unfortunately, this approach is limited in handling all scenarios that can happen in real roads.
Hence, the intelligent braking system should be developed to avoid the accidents in a principled and goal-oriented manner.

  Recently, interest in machine learning has explosively grown up with the rise of parallel computing technology and a large amount of training data. In particular, the success of deep neural network  (DNN)  technique led the researchers to investigate the application of machine learning for autonomous driving.
The  DNN has been applied to autonomous driving from camera-based perception \cite{eedlh,dcnnpd,nnblc} to end-to-end approach which learns mapping from the sensing to the control \cite{eelsc,ddad}.
Reinforcement learning (RL) technique has also been improved significantly as DNN was adopted. The technique, called deep reinforcement learning (DRL), has shown to perform reasonably well for various challenging robotics and control problems.
In \cite{dqn}, the DRL technique called Deep Q-network (DQN) was proposed, which approximates Q-value function using DNN. It was shown that the DQN can outperform human experts
in various Atari video games.
Recently, the DRL is applied to control systems for autonomous driving vehicle in \cite{educn,caccrla}.

%



%


\begin{figure}[thpb]
\centering
\includegraphics[scale = 0.12]{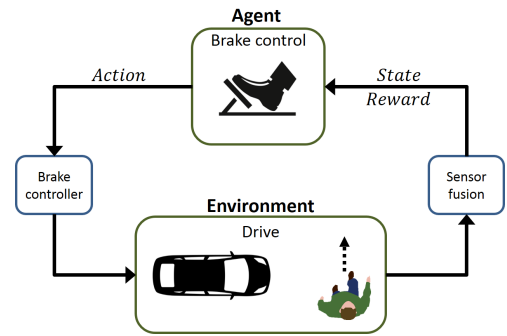}
\caption{ The proposed DRL-based autonomous braking systems.}
\label{fig:overview}
\end{figure}
 In this paper, we propose a new autonomous braking system  based DRL, which can intelligently control the velocity of the vehicle in situations where collision is expected if no action is taken.
 The proposed autonomous braking system is described in Fig. \ref{fig:overview}.
 The agent (vehicle) interacts with the uncertain environment where the position of the obstacle could change in time and thus the risk of collision  at each time step varies as well. 
 The agent receives the information of the obstacle's position using the sensors and adapts the brake control to the state change such that the chance of accident is minimized.

 In our work, we design the autonomous braking system for the urban road scenario where a vehicle faces a pedestrian who crosses the street at a random timing.
 %
 In order to find the desirable brake action for the given pedestrian's location and vehicle's speed, we need to allocate appropriate reward function for each state-action pair.
 %
 In our work, we focus on finding the desirable reward function which  strikes the balance between the penalty imposed to the agent when accident happens and
 the reward obtained when the vehicle quickly gets out of risk.
Using the reward function we carefully designed, we train DQN to learn the policy that decides the timing of brake based on the given pedestrian's state.
We also provide a new DQN design which can rapidly learn the policy to avoid rare accidents.

Via computer simulations, we evaluate the performance of the proposed autonomous braking system. In simulations, we consider the uncertainty of the vehicle's initial velocity, pedestrian's initial position, and whether the pedestrian will cross or not. The experimental results show that the proposed braking system exhibits desirable control behavior for various test scenarios including autonomous emergency braking (AEB) test administrated by Euro NCAP.

The rest of this paper is organized as follows. In Section II, we describe the basic scenarios and the framework of the proposed system. In Section III, we provide the details of the DQN design for autonomous braking.
The experimental results are provided in Section IV and the paper is concluded in Section V.

\begin{figure}[b]
\centering

\includegraphics[scale = 0.18]{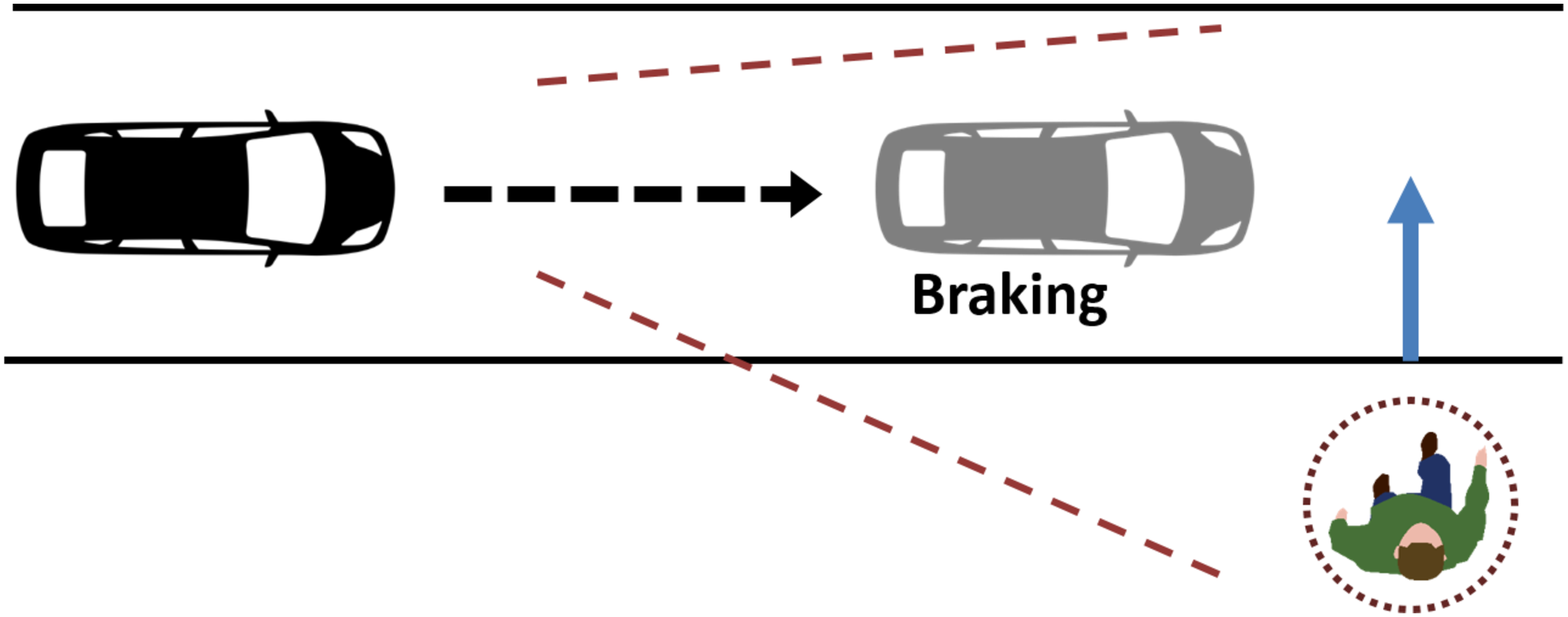}\\(a)\vspace{1cm}

\includegraphics[scale = 0.2]{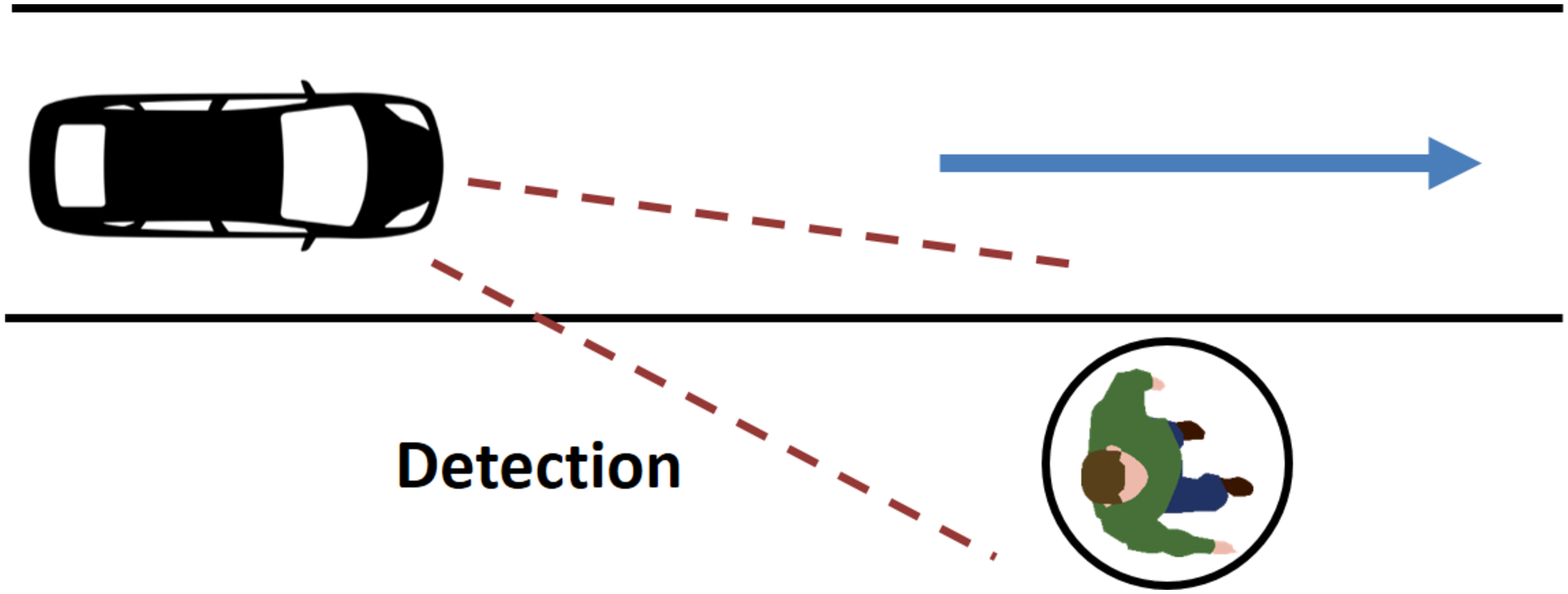}\\(b)
\caption{ Though a pedestrian is detected for the cases (a) and (b), the proper control actions are different for two cases. For case (a), the vehicle should stop in front of the pedestrian while for case (b), the vehicle should be on standby without stepping the brake yet.}
\label{fig:pedestrian}
\end{figure}
\section{System Description}

In this section, we describe the overall structure of the autonomous braking system. We first define the possible scenarios for autonomous braking and explain the detailed operation of the proposed system.

\subsection{Scenarios}
One of the factors that hinders safe driving in autonomous driving is the threat from nearby objects, e.g. pedestrians. Many accidents could happen when the vehicle fails to stop ahead of it when a pedestrian crosses the road. Hence, in order to avoid accidents, the vehicle should detect the threat that can potentially cause accidents in advance and perform appropriate brake actions to stop vehicle in front of the obstacle.  However, there exist various degrees of uncertainty which make the design of autonomous braking challenging such as
 \begin{itemize}
 \item Vehicle's initial velocity
  \item Pedestrian's position
  \item Pedestrian's speed
   \item Pedestrian's crossing timing
  \item Pedestrian's moving direction
  \item Sensor's measurement error
  \item Road's condition
 \end{itemize}
Even if a pedestrian is detected accurately, it is hard to know when it can become a threat to the vehicle. Hence we need appropriate braking strategy for different situations. (see Fig.~\ref{fig:pedestrian}.) That is, for the given state of the pedestrian (i.e. position, velocity), the autonomous braking system should decide what brake action to apply.
  %

\begin{figure}[thpb]
\centering
\includegraphics[scale = 0.09]{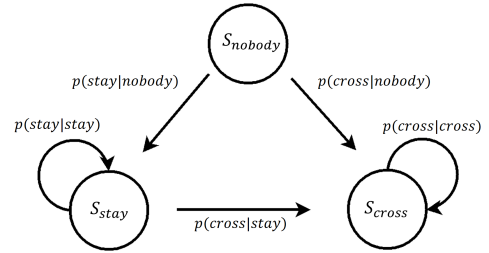}
\caption{Behavior of a pedestrian modeled by the discrete-state Markov process. }
\label{figure:markov}
\end{figure}

In our system, we consider the scenario where behavior of the pedestrian follows the discrete-state Markov process
described in Fig.~\ref{figure:markov}.
The state $S_{nobody}$ implies that the sensors have not detected any obstacle.
Once a pedestrian is detected, the state $S_{nobody}$ can change to the state $S_{stay}$ or the state $S_{cross}$, where $S_{stay}$ is the state that the pedestrian stays at sidewalk and $S_{cross}$ is the state that the pedestrian crosses the road.
The pedestrian's initial position can be either from far-side and near-side of the vehicle  and the pedestrian walking speed can vary between ${v_{ped}}^{min}\ m/s$ and ${v_{ped}}^{max}\  m/s$.
Note that the vehicle's initial velocity is distributed between  ${v_{veh}}^{min}\ m/s$ and ${v_{veh}}^{max}\  m/s$.
In practical scenarios, it is difficult to know the transition probabilities of the Markov process and the distribution of the pedestrian's states. Therefore, reinforcement learning approach can be applied to learn the brake control policy through the interaction with environment.

\subsection{Autonomous Braking System}

\begin{figure}[b]
\centering
\includegraphics[scale = 0.08]{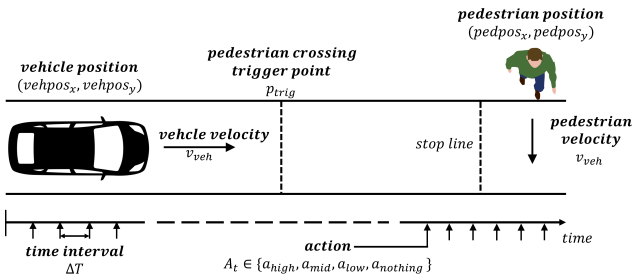}
\caption{Illustration of the autonomous braking operation}
\label{figure:overallsys}
\end{figure}

The detailed operation of the proposed autonomous braking system is depicted in Fig.~\ref{figure:overallsys}.
The vehicle is moving at speed $v_{veh}$ from the position $(vehpos_x, vehpos_y)$.
As soon as  a pedestrian is detected, the autonomous braking system receives the relative position of the pedestrian, i.e.,  $(pedpos_x-vehpos_x, pedpos_y-vehpos_y)$ from the sensor measurements where $(pedpos_x, pedpos_y)$ is the location of the pedestrian.
Using the vehicle's velocity $v_{veh}$ and the relative position $(pedpos_x-vehpos_x, pedpos_y-vehpos_y)$, the vehicle decides whether it will step brake at each time step. The interval between consecutive time steps is given by  $\Delta T$.
We consider four brake actions; no braking ${a_{nothing}}$ and braking $a_{high},a_{mid}$ and $a_{low}$  with different intensities. We can include more brake actions with more refined steps or continuous brake action which are not considered in this work.

\section{Deep Reinforcement Learning for Autonomous Braking System}
In this section, we present the details of the proposed DRL-based autonomous braking system. We first introduce the structure of the DQN and explain the reward function used to train the DQN in details.

\subsection{Structure of DRL}
Our system follows the basic RL structure. The agent performs an action $\it{A_t}$ given state $\it{S_t}$ under policy $\pi$. The agent receives the state as  feedback from the environment and gets the reward $r_t$ for the action taken. The state feedback that the agent takes from sensors consists of the velocity of the vehicle ${v_{veh}}$ and the relative position to the pedestrian, $(pedpos_x-vehpos_x, pedpos_y-vehpos_y)$ for the past $n$ time steps.  Possible action that agent can choose is among deceleration ${a_{high},a_{mid},a_{low}}$ and keeping the current speed ${a_{nothing}}$. The goal of our proposed autonomous braking system is to maximize the expected accumulated reward called ``value function" that will be received in the future within an episode.
 Using the simulations,  the agent learns from interaction with environment episode-by-episode.  One episode starts when a pedestrian is detected.
Note that the initial position of the pedestrian and the initial velocity of the vehicle are random. The vehicle drives on a straight way based on the brake policy ${\pi}$. If the distance between the vehicle and the pedestrian is less than the safety distance $l$, it is considered as a collision event. (see Fig.~\ref{figure:overallsys}.)  The episode ends if at least one of the following events occurs
\begin{itemize}
\item $Stop$ : the vehicle completely stops, i.e.,  $v_{veh} = 0$.
\item $Bump$ : the vehicle passes the safety line $l$ when the pedestrian is crossing road.
\item $Pass$ : the vehicle passes the pedestrian without accident.
\item $Cross$ : the pedestrian completely crosses the road and reaches the opposite side.

\end{itemize}
Once one episode ends, the next episode starts with the state of environment and the value function reset.

\subsection{Deep Q-Network}
 Q-learning is one of the popular RL methods which searches for the optimal policy in an iterative fashion \cite{sutton}. Basically, the Q-value function $q_\pi(s,a)$ is defined as
 \begin{equation}
q_\pi(s,a) = \mathbb{E}_\pi[\Sigma_{k = 0}^{\infty} \gamma^k r_{t+k+1}|S_t = s, A_t = a]
\end{equation}
for the given state $s$ and action $a$, where $r_t$ is the reward received at the time step t.
The Q-value function is the expected sum of  the future rewards which
 indicates how good the action $a$ is given the state $s$ under the policy of the agent $\pi$.
 The contribution to the Q-value function decays exponentially with the discounting factor $\gamma$ for the rewards with far-off future. For the given Q-value function, the greedy policy  is obtained as
\begin{align} \label{eq:policy}
\pi(s) = \arg \max_{a} q_\pi(s,a).
\end{align}
One can show that for  the policy in (\ref{eq:policy}), the following Bellman equation should hold \cite{sutton};
\begin{equation}
q_{\pi}(s,a) = \mathbb{E}\left[r_{t+1} + \gamma \max_{a'}q_{\pi}(S_{t+1},a')|S_t = s, A_t = a \right].
\end{equation}
In practice, since it is hard to obtain the exact value of $q_{\pi}(s,a)$ satisfying the Bellman equation, the Q-learning method
uses the following update rule for the given one step backups $S_t$, $A_t$, $r_{t+1}$, $S_{t+1}$;
\begin{align}
q_{\pi}(S_t,A_t )& \leftarrow q_{\pi}(S_t,A_t )  \nonumber \\
&+ \alpha \left(r_{t+1} +\gamma \max_a q_{\pi}(S_{t+1},a) - q_{\pi}(S_t,A_t ) \right)
\end{align}
However, when the state space is continuous, it is impossible to find the optimal value of the state-action pair ${q_*(s,a)}$ for all possible states.
To deal with this problem, the DQN method was proposed, which approximates the state-action value function ${q(s,a)}$ using the DNN, i.e., ${q(s,a) \approx {q}_\theta(s,a)}$ where $\theta$ is the parameter of the DNN \cite{dqn}.
 The parameter $\theta$ of the DNN is then optimized to minimize the squared value of
the  temporal difference error $\delta_t$
\begin{equation}
\delta_t =  r_{t+1} + \gamma \max_{a'} q_{\theta}(S_{t+1},a')  - q_\theta(S_t,A_t)
\label{bellmandif}
\end{equation}
For better convergence of the DQN, instead of estimating both $q(S_{t},A_t)$ and $q(S_{t+1},a')$ in (\ref{bellmandif}), we approximate $q(S_{t},A_t)$ and $q(S_{t+1},a')$ using the Q-network and the target network parameterized by $\theta$ and $\theta^-$, respectively \cite{dqn}. The update of the target network parameter $\theta^-$ is done by cloning Q-network parameter $\theta$, periodically. Thus, (\ref{bellmandif}) becomes
\begin{align}
\delta_t & = r_{t+1} + \gamma \max_{a'}  q_{\theta^-}(S_{t+1},a')  - q_\theta(S_t,A_t)
\end{align}
To speed up convergence further, replay memory  is adopted to store a bunch of one step backups and use a part of them chosen randomly from the memory  by batch size \cite{dqn}. The backups in the batch is used to calculate the loss function $L$ which is given by
\begin{align}
L= \Sigma_{t \in B_{replay}}{\delta_t}^2,
\end{align}
where $B_{replay}$ is the backups in the batch selected from replay memory.
Note that the optimization of parameter $\theta$ for minimizing the loss $L$ is done through the stochastic gradient decent method.

\subsection{Reward Function}
Unlike video games,  the reward should be appropriately  defined by a system designer in autonomous braking system.  As mentioned, the reward function determines the behavior of the brake control. Hence, in order to ensure the reliability  of the brake control, it is crucial to use the properly defined reward function. In our model, there is conflict between two intuitive objectives for  brake control; 1) collision should be avoided no matter what happens and 2)  the vehicle should get out of the risky situation quickly. If it is unbalanced, the agent becomes either too conservative or reckless.  Therefore, we should use the reward function which balances two conflicting objectives.
%
Taking this into consideration, we propose the following reward function

\begin{align} \label{eq:rt}
r_t = -(\alpha(pedpos_x-&vehpos_x)^2 +\beta)decel \nonumber \\
 - (\eta{v_t}^2+\lambda)& \textbf{1}(S_{t} = bump) \\
 \alpha,\beta,\eta,&\lambda > 0 \nonumber
\end{align}
where $v_t$ is the velocity of the vehicle at the time step $t$, $decel$ is difference between $v_t$ and $v_{t-1}$ and $\textbf{1}(x = y)$ has a value of $1$ if the statement inside is true and $0$ otherwise.
The first term $ -(\alpha(pedpos_x-vehpos_x)^2 +\beta)decel$ in the reward function prevents the agent from braking too early by giving penalty proportional to squared distance between the vehicle and pedestrian. It guides the vehicle to drive without deceleration if the pedestrian is far from the vehicle.
On the other hand, the term $-(\eta {v_t}^2 + \lambda) \textbf{1}(S_{t} = bump)$ indicates the penalty that the agent receives when the accident occurs. Note that this penalty is a function of the vehicle's velocity, which reflects the severe damage to the pedestrian in case of high velocity at collision. Without such dependency on the velocity, the agent would not reduce the speed  in situation when the accident is not avoidable.
The constants $\alpha$, $\beta$, $\eta$ and $\lambda$ are the weight parameters that controls the trade-off between two objectives.

\subsection{Trauma Memory}

As mentioned in the previous section, autonomous braking systems should learn both of the conflicting objectives. However, when we train the DQN with the reward function in (\ref{eq:rt}), we find that the learning performance is not stable since collision events rarely happen and thus there remains only a few one-step backups associated with the collisions in the replay memory.
As a result, the probability of picking such one-step backups is small and the DQN does not have enough chance to learn to avoid accidents in practical learning stage.
To solve this issue, we propose so called ''trauma" memory which is used to store only the one-step backups for the rare events (e.g., collision events in our scenario). 
While the one step backups are randomly picked from the replay memory, some fixed number of backups associated with the collision events are randomly selected from the trauma memory and used for training together.
In other words, with the trauma memory, the loss function $L$  is modified to
\begin{align}
L = \Sigma_{t \in B_{replay}}{\delta_t}^2 + \Sigma_{t \in B_{trauma}}{\delta_t}^2
\end{align}
where $B_{trauma}$ is the backups randomly picked from trauma memory.
Trauma memory persistently reminds the agent of the memory on the accidents regardless of the current policy, thus allowing the agent to learn to maintain speed and avoid collisions reliably.

\section{Experiments}
In this section, we evaluate the performance of the proposed autonomous braking system via computer simulations.
\subsection{Simulation Setup}

In simulations, we used the commercial software \emph{PreScan} which models vehicle dynamics in real time \cite{prescan}.
We generated the environment in order to train the DQN by simulating the random behavior of the pedestrian.
In the simulations, we assume that the relative location of the pedestrian is provided to the agent. To make the system practical, we add slight measurement noise to it.
In each episode, the initial position of vehicle is set to $(0,0)$.
Time-to-collision $TTC$ is chosen according to the uniform distrubution between $1.5 \ s$ and $4 \ s$.
The initial velocity of the vehicle is uniformly distributed between ${v_{init}}^{min}=2.78\ m/s \ (10 \ km/s)$ and ${v_{init}}^{max} = 16.67\ m/s \ (60 \ km/h)$.
At the beginning of the episodes, the position of the pedestrian is fixed to $5 * v_{init} $ meters away from the position of the vehicle. The pedestrian stands either at the far-side or at near-side of the vehicle with equal probability.
The behavior of the pedestrian follows one of two scenarios below;
\begin{itemize}
\item Scenario 1 : Cross the road
\item Scenario 2 : Stay at initial position.
\end{itemize}
During training, either of two scenarios is selected with equal probability.
In Scenario 1, the pedestrian starts to move when the vehicle is crossing at the ``pedestrian crossing point" $p_{trig}=(5-TTC)*v_{init}$. (see Fig.~\ref{figure:overallsys}.) 
 The safety distance $l$ for the pedestrian is  set to $3$ m.
 The agent chooses the brake control among  $a_{high} = -9.8\ m/s^2$, $a_{mid} = -5.9\ m/s^2$, $a_{high} = -2.9\ m/s^2$ and $a_{nothing} = 0\ m/s^2$
 every  $\Delta T = 0.1$ second.
 The detailed simulation setup is summarized below.

\begin{itemize}
\item Initial velocity of vehicle $ v_{init} \sim U(2.78,16.67)\ m/s$
\item Velocity of pedestrian $v_{ped} \sim U(2,4) \ m/s$
\item Time-to-collision $TTC \sim U(1.5,4) \ s$
\item Initial pedestrian position $ pedpos_x = 5 * v_{init}\ m$
\item Trigger point \ $ p_{trig} = (5 - TTC)*v_{init} \ m$
\item Safety line $ l = 3 \ m$
\item $\Delta T = 0.1 \ s$
\item $ a_{high}, a_{mid}, a_{low}, a_{nothing} = \{ -9.8, -5.9, -2.9, 0\}\ m/s^2$

\end{itemize}

\subsection{Training of DQN}

\begin{figure}[thpb]

\centering
\includegraphics[scale = 0.4]{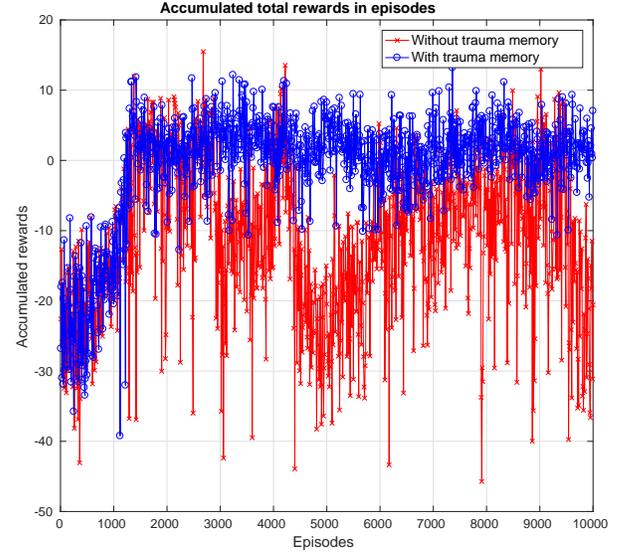}
\caption{ Achieved value function achieved during training}
\label{figure:learningcurve}
\end{figure}

The neural network used for the DQN  consists of the fully-connected layers with five hidden layers. RMSProp algorithm  \cite{rmsprop} is used to minimize the loss with learning rate $\mu = 0.0005$. The number of position data samples used as a state is set to $n=5$.  We set the size of the replay memory to 10,000 and that of the trauma memory to 1,000. We set the replay batch size to 32 and trauma batch size to 10.  The summary of the DQN configurations used for our experiments is provided below;
\begin{itemize}
\item State buffer size: $n = 5$
\item Network architecture: fully-connected feedforwared network
\item Nonlinear function: leaky ReLU \cite{relu}
\item Number of nodes for each layers : [15(Input layer), 100, 70, 50, 70, 100, 4(Output layer)]
\item RMSProp optimizer with learning rate 0.0005 \cite{rmsprop} 
\item Replay memory size: 10,000
\item Replay batch size: 32
\item Trauma memory size: 1,000
\item Trauma batch size = 10
\item Reward function: $\alpha = 0.001,\ \beta = 0.1,\ \eta = 0.01,\ \lambda = 100$
\end{itemize}
Fig. ~\ref{figure:learningcurve} provides the plot of the total accumulated rewards i.e., value function achieved  for each episode when training is conducted with and without trauma memory.  We observe that with trauma memory the value function converges after 2,000 episodes and high total reward is steadily attained after convergence while without trauma memory the policy does not converge and keeps fluctuating.

\subsection{Test Results}
\begin{figure}[thpb]
\centering
\includegraphics[scale = 0.4]{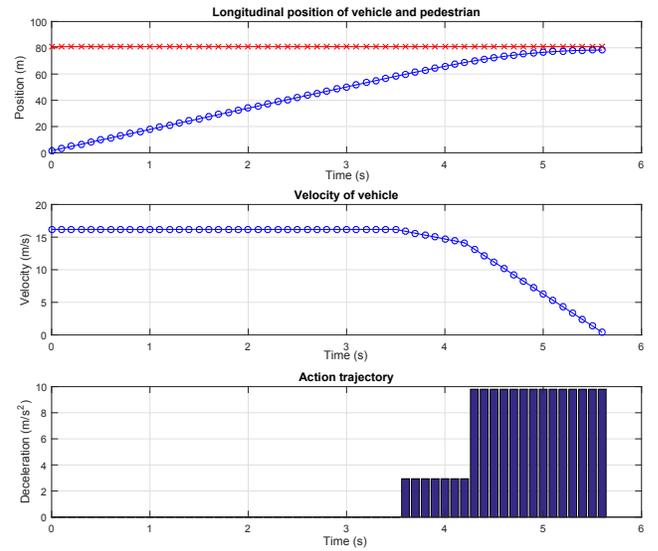}
\caption{Trajectory of position, velocity and actions in one example episode for the case $TTC = 1.5s$}
\label{figure:action_traj}
\end{figure}

Safety test was conducted for several different $TTC$ values. Collision rate is measured for 10,000 trials for each TTC value.
\begin{table*}[t]
\caption{Collision rate in test scenarios}
\begin{center}
\begin{tabular}{|c||c|c|c|c|c|c|c|c|c|c|c|c|c|c|c|c|c|c|}
\hline
$TTC$ (s) & 0.9  & 1.1  & 1.3  & 1.5  & 1.7  & 1.9  & 2.1  & 2.3  & 2.5 & 2.7 & 2.9  & 3.1  & 3.3 & 3.5 & 3.7 & 3.9  \\ 
\hline
Collision rate (\%) &61.29 & 18.85  &0.74 & 0  & 0 & 0  & 0  & 0   & 0  & 0  & 0 & 0  & 0  & 0   & 0  &0     \\
\hline

\end{tabular}

\end{center}
\end{table*}
Table \RN{1} provides the collision rate for each $TTC$ value for the test performed for Scenario 1. The agent avoids collision successfully for $TTC$ values above 1.5s. For the cases with $TTC$ values less than 1.5s, we observe some collisions. According to our analysis on the trajectory of braking actions, these are the cases where collision was not avoidable due to the high initial velocity of the vehicle even though full braking actions were applied.  
The agent passed the pedestrian without unnecessary stop for all cases in the Scenario 2. The detailed trajectory of the brake actions for one example case is shown in Fig. ~\ref{figure:action_traj}. 
Fig.~\ref{figure:action_traj} (a) shows the trajectory of the position of the vehicle and the pedestrian recorded every $0.1$ s. The velocity of the agent and the brake actions applied are shown in Fig.~\ref{figure:action_traj} (b) and  (c), respectively. The vehicle starts to decelerate about 20 m away from the pedestrian and completely stops about 5 m ahead, thereby accomplishing collision avoidance. We observe that  weak braking actions are applied in the beginning part of deceleration and then strong braking actions come as the agent gets close to the pedestrian.



\begin{figure}[thpb]
\centering
\includegraphics[scale = 0.35]{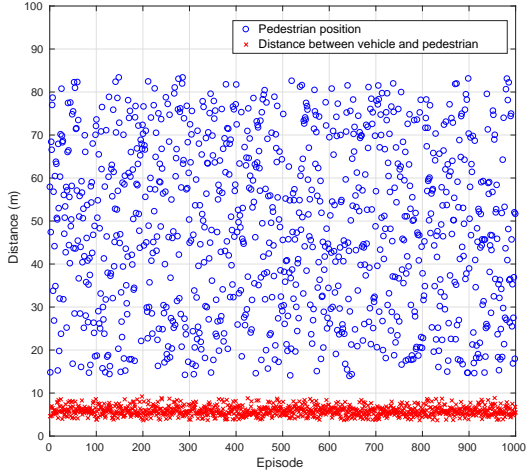}
\caption{Initial position of the pedestrian and relative distance between the pedestrian and vehicle after the episode ends.}
\label{figure:distresult}
\end{figure}

Fig. ~\ref{figure:distresult} shows how the initial position of the pedestrian and the relative distance between the pedestrian and vehicle are distributed for 1,000 trials in the scenario 1. We see that the vehicle stops around $5\ m$ in front of the pedestrian for most of cases. This seems to be reasonable safe braking operation  considering the safety distance of $l=3$ m. Note that this distance can be adjusted by changing the reward parameters. Overall, the experimental results show that the proposed autonomous braking system exhibits consistent brake control performance for all cases considered.

\subsection{Test Results for Euro NCAP AEB Pedestrian Test}
Additional autonomous emergency braking (AEB) pedestrian tests are conducted. We follow the test procedure specified by Euro NCAP test protocol \cite{EURONCAP}, \cite{Protocol}. Tests are conducted for both farside (CVFA test) and nearside (CVNA test) under the velocity range between 20 to 60 $km/h$ with 5 $km/h$ interval. $TTC$ is set to 4 $s$ and the pedestrian crosses the road at 8 $km/h$ for CVFA and 5 $km/h$ for CVNA. Tests are scored according to the rating parameters and the metric suggested in \cite{Protocol}. The proposed system passed all tests without collision and the rating scores acquired by the proposed method are shown in Table \RN{2}.

\begin{table}[h]
\caption{AEB test result}
\begin{center}
\begin{tabular}{|c||c|c|c|c|c|c|c|c|c|}
\hline
$v_{init} (km/s)$ & 20  & 25  & 30  & 35 & 40 & 45 & 50  & 55 &  60  \\
\hline
$CVFA \ score$ & 1 & 2 & 2 & 3 & 3 & 3 & 2 & 1 & 1 \\
\hline
$CVNA \ score$ & 1 & 2 & 2 & 3 & 3 & 3 & 2 & 1 & 1 \\
\hline
\end{tabular}
\end{center}
\end{table}

\section{CONCLUSIONS}
We have presented the new autonomous braking system based on the deep reinforcement learning. The proposed system learns an intelligent way of brake control from the experiences obtained under the simulated environment.
We designed  the autonomous braking systems  using the DQN method with carefully designed reward function and enhanced stability of learning process by modifying the structure of the DQN. We showed through computer simulations that
the proposed autonomous braking system exhibits desirable and consistent brake control behavior for various scenarios where behavior of the pedestrian is uncertain.

\end{document}